\documentclass[conference]{IEEEtran}
\IEEEoverridecommandlockouts               

\usepackage{cite}
\usepackage{amsmath,amssymb,amsfonts}
\usepackage{algorithmic}
\usepackage{graphicx}
\usepackage{textcomp,xcolor}
\usepackage{textcomp,xcolor}

\usepackage{titlesec}
\renewcommand\thesubsubsection{\arabic{subsubsection}} 
\titleformat{\subsubsection}
    {\normalfont} 
    {\thesubsubsection.} 
    {0.5em} 
    {} 
\titlespacing*{\subsubsection}{0pt}{2.5ex plus 1ex minus .2ex}{1ex plus .2ex}

\usepackage[intoc]{nomencl}  
\makenomenclature
\usepackage{siunitx}         
\usepackage{tabularx}
\usepackage[draft=true]{hyperref}
\usepackage{array} 
\usepackage{subcaption}      
\usepackage{float}
\usepackage{booktabs} 
\usepackage{threeparttable}

\begin{document}
\title{CANSURF: An ASV-View Can Dataset and Benchmark for Detection and Tracking of Surface-Level Debris%

\thanks{Z. Aljundi, A. Moosa, M. Elemam, and Z. F. Rahmatullah, ``CANSURF: An ASV-View Can Dataset and Benchmark for Detection and Tracking of Surface-Level Debris,'' \textit{2025 8th International Conference on Signal Processing and Information Security (ICSPIS)}, pp. 1--6, 2025. DOI: \href{https://doi.org/10.1109/ICSPIS67605.2025.11318414}{10.1109/ICSPIS67605.2025.11318414}}}

\author{%
  \IEEEauthorblockN{Zaid Aljundi}
  \IEEEauthorblockA{\textit{School of Mathematical and Computer Sciences}\\
                    Heriot-Watt University Dubai\\
                    Dubai, United Arab Emirates\\
                    zaidaljundi05@gmail.com}
  
  \vspace{1.5ex} 

  \IEEEauthorblockN{Mostafa Elemam}
  \IEEEauthorblockA{\textit{School of Engineering and Physical Sciences}\\
                    Heriot-Watt University Dubai\\
                    Dubai, United Arab Emirates\\
                    Mostafa.w.elemam16@gmail.com}

  \and 

  \IEEEauthorblockN{Abdullah Moosa}
  \IEEEauthorblockA{\textit{School of Engineering and Physical Sciences}\\
                    Heriot-Watt University Dubai\\
                    Dubai, United Arab Emirates\\
                    abdullah.zmoosa10@gmail.com}

  \vspace{1.5ex} 

  \IEEEauthorblockN{Zahra F. Rahmatullah}
  \IEEEauthorblockA{\textit{School of Mathematical and Computer Sciences}\\
                    Heriot-Watt University Dubai\\
                    Dubai, United Arab Emirates\\
                    Itzimbey@gmail.com}
}

\maketitle
\bstctlcite{IEEEexample:BSTcontrol}
\IEEEpubid{\makebox[\columnwidth]{979-8-3315-8529-7/25/\$31.00~\copyright2025 IEEE \hfill} \hspace{\columnsep}\makebox[\columnwidth]{ }}
\IEEEpubidadjcol


\begin{abstract}
Surface-level marine debris remains a practical bottleneck for autonomous clean-up, where small, reflective targets (e.g., aluminum cans) must be detected at distance under glare, ripples, and partial submersion. This paper presents, an ASV vision system and a new surface-can dataset. The dataset comprises $\sim$7.3k raw images extracted from videos and annotated with bounding boxes, expanded via ten augmentation types to $\sim$57k training/validation images spanning diverse lighting and water states. A family of detector and detector–tracker pipelines tailored to surface operations were benchmarked. Training YOLOv11 on CANSURF boosts performance 12x over generic datasets, highlighting the dataset's value. Experiments show that YOLOv11+ByteTrack yields the most stable tracks (fewer identity switches) and stronger multi-object accuracy under, while YOLOv11+SAHI increases recall on far-field cans at the cost of lower precision in full-context inputs. Given the  mission profile, single-can pickup with approach and grab, YOLOv11 + SAHI proves better for detecting the maximum number of cans. No prior open dataset targets aluminum cans on water from a surface-level viewpoint; this dataset fills this gap and supports reproducible evaluation.

\end{abstract}

\vspace{0.04cm}

\begin{IEEEkeywords}
Autonomous Surface Vehicles (ASV), Marine Debris Detection, Object Detection, Benchmark Datasets, Small Object Detection, Multi-Object Tracking, Domain Adaptation, Real-time systems.
\end{IEEEkeywords}

\section{Introduction}
\label{sec:intro}

Marine litter on the water surface poses operational and ecological risks in harbors, canals, and coastal zonesBeverage containers, including aluminum cans, are among the most frequently recorded items in coastal cleanups and are listed in marine debris response programs such as the NOAA Marine Debris Program and the Ocean Conservancy Program. ~\cite{noaa_handling_2020, oc_2020_bythenumbers}. Although recent work has advanced debris detection from satellites and underwater platforms, there remains a practical gap in open, surface-level datasets that reflect the visual challenges faced by autonomous surface vehicles (ASV) in real-world conditions (sunglint, waves/foam, partial submersion, small apparent object size)~\cite{marida_plosone_2022}.

\begin{figure}[H]
    \centering
    \includegraphics[width=1\linewidth]{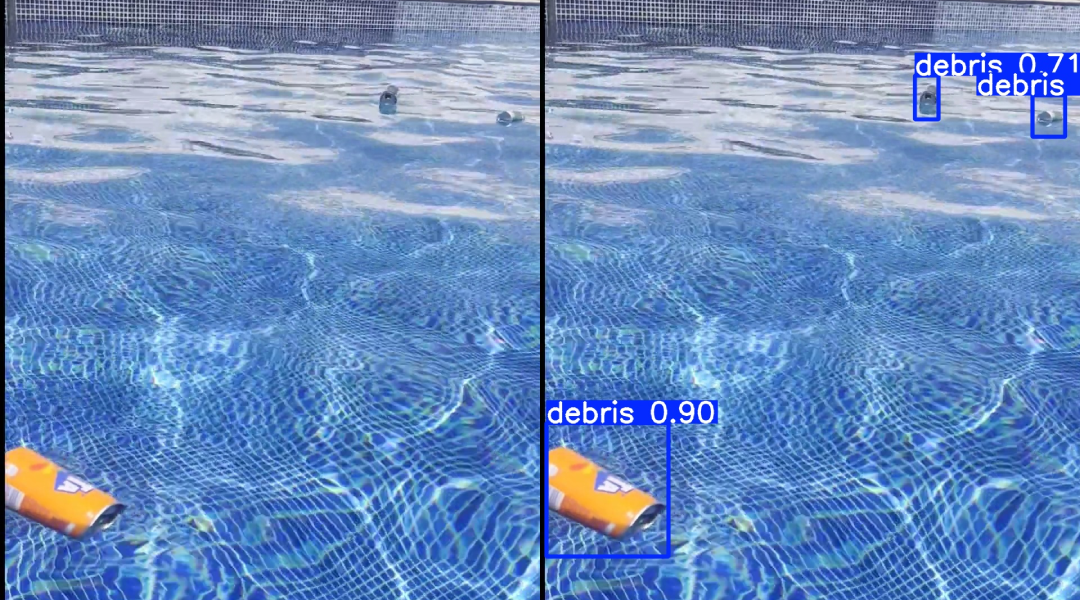}
    \caption{Example tracking results from the top-performing YOLOv11 + SAHI system.}
    \label{fig:validation_image}
\end{figure}

\par With the mission of collecting floating debris, consisting of aluminum cans, this paper makes two contributions:
\begin{enumerate}
\item An open dataset of aluminum cans floating on the water surface, collected from an ASV perspective. The dataset includes cans of different colors and sizes under varying illumination, with annotations for detection and tracking. Unlike existing datasets that focus on \emph{underwater} (e.g., Trash-ICRA19 ~\cite{trashicra19_repo}, TrashCan ~\cite{trashcan_drum}) or \emph{satellite} imagery (e.g., MARIDA ~\cite{marida_plosone_2022}), this dataset targets near-surface, close-range conditions relevant to ASVs.
    
    \item A vision pipeline combining a modern object detector with Slicing Aided Hyper Inference (SAHI) ~\cite{sahi_arxiv} for improved recall on small, distant targets, and ByteTrack for maintaining track continuity under occlusion and submersion. SAHI tiles frames at inference time to boost small-object sensitivity without retraining, while Bytetrack uses appearance cues to reduce identity switches—both critical for surface scenes where cans are small, intermittently obscured, and drift with currents.
\end{enumerate}
\section{literature review}
Understanding the alternative approaches to the use of object detection in autonomous surface vehicles is crucial to map the benefits and detriments for the purpose of providing different systems to choose from based on specific application needs. 

Some of the main concerns of using the available object detection models through autonomous surface vehicles include noise and difficulty in identifying objects whose dimensions are relatively small compared to the image's dimensions.

Among the existing models, a modified version of VarifocalNet- which uses the ResNet50 network as its backbone CNN, was used to test and address these problems. The model utilised a feature pyramid network, similar to the SPPF block of YOLOv11, to build on already extracted features and create a feature pyramid for the purpose of increasing the model’s invariance in detection target objects of multiple dimensions within the image. The feature maps then go through a deformable convolutional block which helps detect objects of irregular shape or with obstructions that might present only partial views of the target object by using a deformable kernel instead of a fixed size kernel. In testing, the authors found an average precision rate of 78.9\% with the VarifocalNet detection algorithm examined in the study, which is an improvement over the precision rates of YOLOv3, Faster R-CNN and Cascade R-CNN by about 10-25\% on average~\cite{jmse10111783}.A large point of criticality with the use of this model was its trouble in detecting items that appear small relative to the frame’s size.

The work done in different papers presents promise for more diverse datasets to handle detection of water pollution and an ideal dataset for general debris detection on water surfaces with an optimised model. Moving on to datasets, there exists a dataset called “PoTATO” made for the purpose of plastic bottle detection in natural bodies of water- which is the the most similar existing dataset ~\cite{batista2024potatodatasetanalyzingpolarimetric}. On top of the object detection models being used in the paper, the authors present a technique in which light polarisation is used to improve detection accuracy by using polarisation information extracted from the image to increase the visual construct of partially obstructed/hidden target objects. The dataset includes images captured directly from the host ASV’s point of view over different weather conditions in a lake. The study found that training on a color with diffusal of light channel performed best out of any channel for general detection, combining the effectiveness of accurate detection of smaller target objects using RGB channels in training, and the effectiveness of accurate detection of larger objects using degree of light polarization channels in training. The authors claim to have experimented with YOLOv5 on one of the pre-existing general datasets that were used on the “bottles” class and have found poor generalization and invariance, performing poorly in the detection of smaller objects. The authors also fine-tuned three models, particularly YOLOv5, Faster R-CNN, and RetinaNet, using the training set and observed an improvement in performance, with Faster R-CNN yielding the highest precision rate.

The paper identifies the detection of larger objects as a strength and claims that the detection of smaller objects continues to remain a challenge in the detection of water pollution, which the proposed methodology of using SAHI may potentially address. To reach an ideal dataset for general water debris detection, multiple sub-datasets for different types of water debris can be used to train an ideal model, which can be exemplified when combining datasets like PoTATO which focuses on plastic debris with CANSURF.

A paper titled “Construction of a Real-Time Detection for Floating Plastics in a Stream Using Video Cameras and Deep Learning”  delves into the detection and tracking of plastic debris in water streams through using  slightly modified YOLOv8 nano model for object detection and the Deep-sort algorithm for object tracking. Trained on a 4,162 image dataset at 1000 epochs, the trained model showcased an impressive mAP at IoU=0.5 with a value of .99 and .992 in the validation and testing steps~\cite{s25072225}.
\vspace{-0.025cm}

However, the tracking portion of the paper showed less success using the Deep-sort algorithm. The deep-sort algorithm, which uses a Kalman filter to predict the next locations of the debris and the Hungarian algorithm to compare the predicted location to the newly detected location for updating the tracked path, was only able to successfully track 6 units of plastic debris out of 32 consistently from the start of the unit’s path until it’s set finish line. The tracking either failed completely on the other 26 units, or acted inconsistently as the plastic debris kept being obstructed from the view of the camera due to constant submersion into the stream, in which they were treated as newly identified pieces of debris after their re-emergance. A possible limitation mentioned in the paper is the factor of dynamic water and weather conditions such as heavy rainfall. The poor performance from the deep-sort algorithm proposed the potential effectiveness in less dynamic environments, like still water surfaces where a USV might operate unlike a flowing stream, and on items that are not transparent like coloured aluminum cans whose surfaces contrast with water surfaces where they float, reducing constant new ID assignment, making its use worth benchmarking against other algorithms such as Bytetrack, which is one of the state-of-the-art algorithms achieving the highest metrics in the public detector dataset for object tracking\cite{zhang2022bytetrackmultiobjecttrackingassociating}.

\label{sec:lit}
\section{methodology}
\subsection {Dataset}
\subsubsection{Dataset Compilation}
The dataset, consisting of 57,012 images divided with 55,246 images in the training subdirectory (96\%) and 1,776 images in the validation subdirectory (4\%) consists of a mixture of original images captured and a partition of images collected from various Roboflow datasets ~\cite{o_20bg_2_dataset} ~\cite{canettes-wjjyb_dataset} featuring aluminum cans including augmented images. The unusually large proportion is due to explicitly applying the augmentations to the training subdirectory as the validation subdirectory remains free from augmentation to preserve integrity of data in validating performance. The total number of images in the dataset excluding augmentations is 7,072, with 4,069 images (57\%) being sourced from the roboflow datasets used, and 3,003 (43\%) images captured originally from the authors, with approximately 5.3k images (75\%) in the training subdirectory and approximately 1.7k (25\%) in the validation subdirectory. The process of image collection consisted of capturing videos of various aluminum cans at 30 FPS floating in different accessible bodies of water to the author, being mostly in the form of personal swimming pools, after which, a Python script was used to convert every 15th frame of the video into an image. The large gap in frames came due over-fitting in previous tests that reduced the generalization of the model and, therefore, led to poor performances. The images featured one to ten cans each at varying distances to the camera, ranging from close-ups that took the view of the entire image to distant views where the cans covered only a few pixels each. After splitting the videos into frames, a script was used to organise via a random split protocol according to the 75/25 ratio into a dataset, and augmentations were applied to the training subdirectory. Although overlapping of sequential frames of the same video appears in both training and validation subdirectories, the combination of random distribution as well as the fact that most videos were of 10-30 seconds length, with only a few being over a minute, offering more diverse frames and perspectives, mitigates the problem. Table I showcases a data card that summarises some of features of the dataset.

{ 
\setlength{\intextsep}{10pt} 

\begin{table}[h!]
\centering
\caption{Dataset Characteristics}
\label{tab:dataset_chars}
\begin{tabular}{@{}ll@{}} 
\hline
\textbf{Characteristic} & \textbf{Value} \\
\hline
Lighting & Natural sunlight, {\fontencoding{T1}\fontfamily{lmr}\selectfont\textasciitilde} 20k lum \\
Frames & 7{,}072 \\
FPS & 30 \\
Resolution & 1280x720, 1080x1920 \\
Camera & OV5693 80 degree FF \\
Sites & Swimming pools \\
Can Dist. (m) & 0--13 \\
\hline
\end{tabular}
\end{table}
}

The authors then used a python script to manually annotate every frame by hand by dragging tightly fitted bounding boxes around each can. Edge cases were handled by annotating a bounding box that extended to the edge of the frame. Occluded cans were labelled such that the bounding box extended to include the full can volume if the majority of the can was visible, otherwise, capturing only visible features. No formal inter-annotator agreement was used as annotators annotated their own set of unique frames. However, before training the images and annotations were verified for their quality, making sure that no annotations were out of the frame.

\subsubsection{Data Augmentation}
\begin{figure}[h!] 
    \centering
    
    \begin{subfigure}[b]{3cm}
        \centering
        \includegraphics[width=3cm, height=3cm]{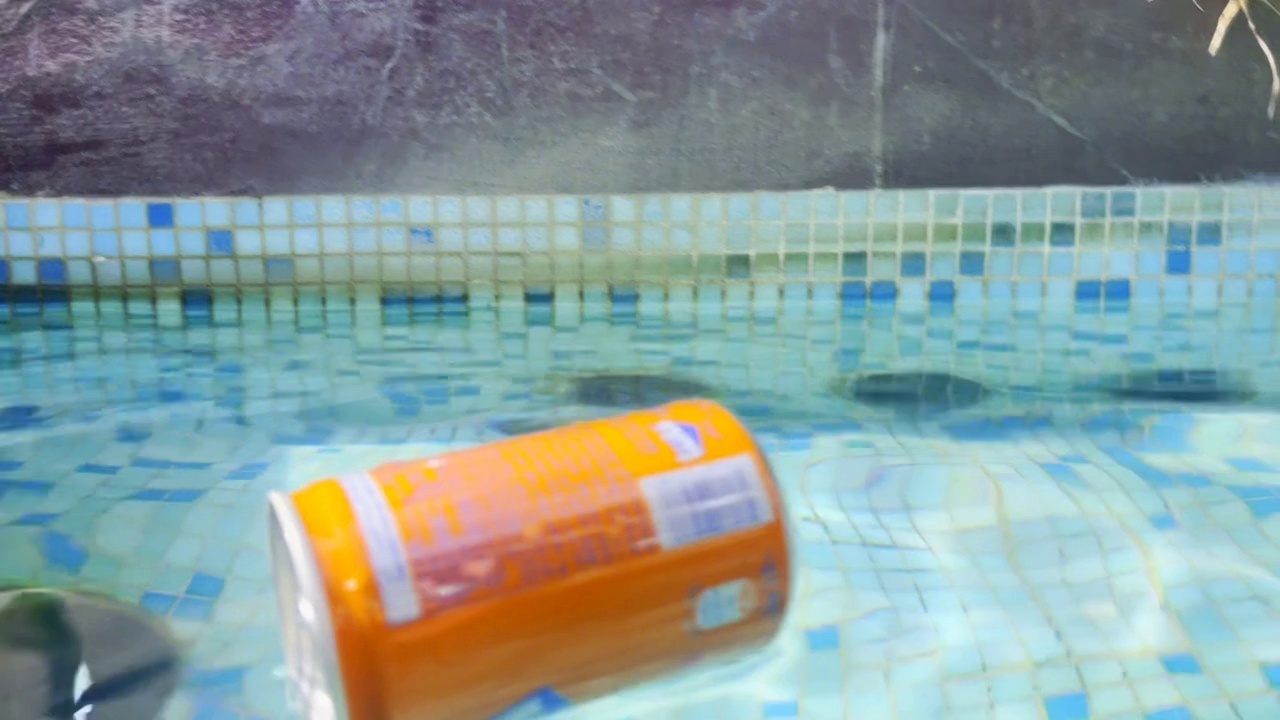} 
        \caption{Original frame}
        \label{fig:image1}
    \end{subfigure}
    \hspace{0.3cm} 
    \begin{subfigure}[b]{3cm}
        \centering
        \includegraphics[width=3cm, height=3cm]{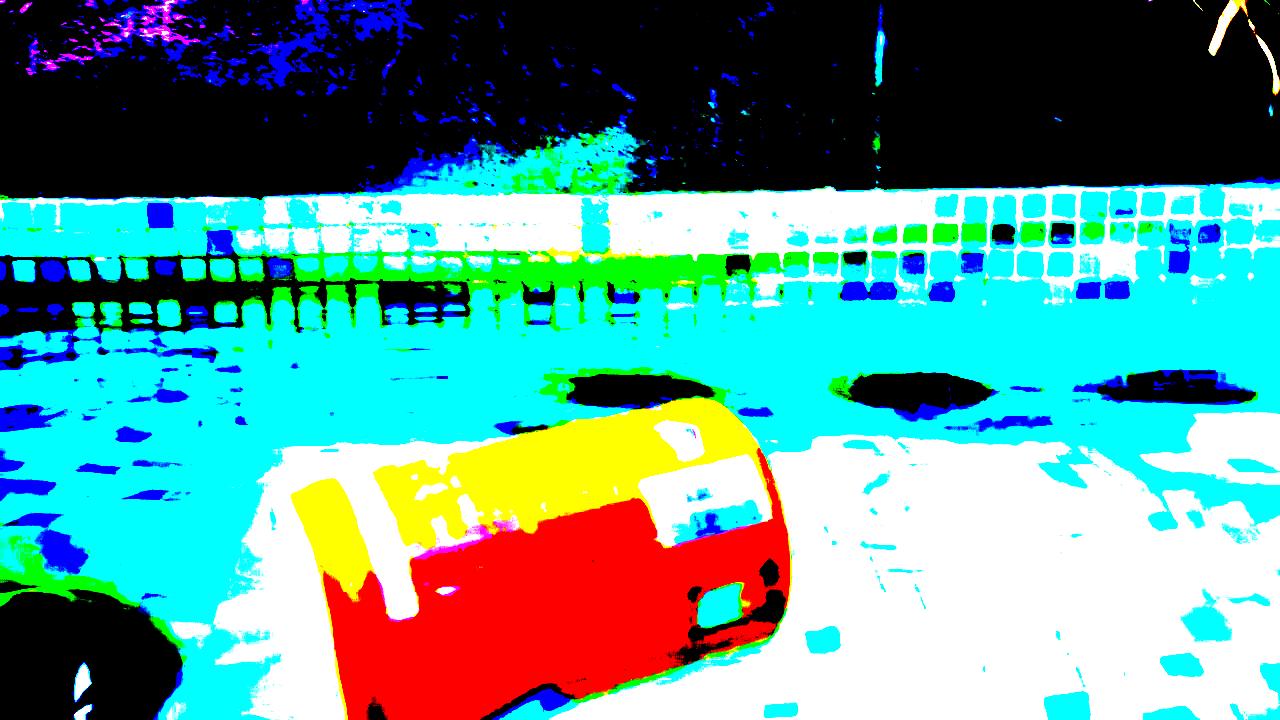} 
        \caption{Color augmented}
        \label{fig:image2}
    \end{subfigure}

    \vspace{0.3cm} 

    \begin{subfigure}[b]{3cm}
        \centering
        \includegraphics[width=3cm, height=3cm]{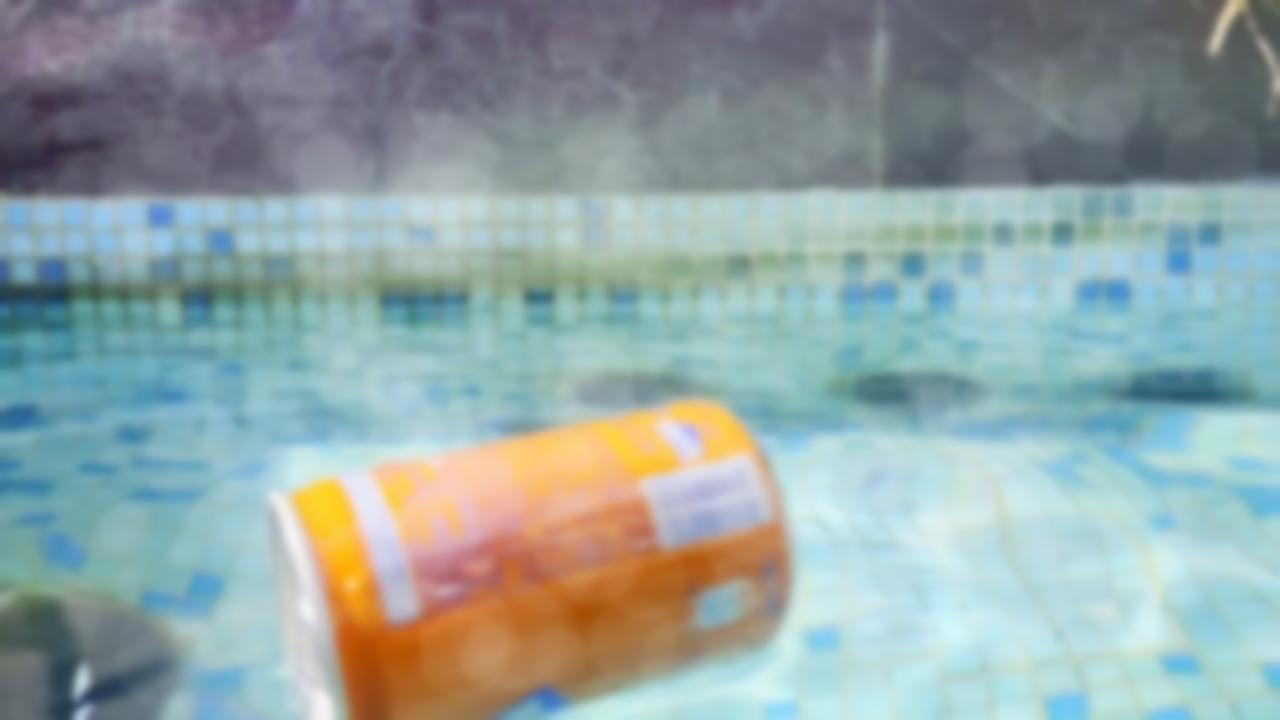} 
        \caption{Weather augmented}
        \label{fig:image3}
    \end{subfigure}
    \hspace{0.3cm} 
    \begin{subfigure}[b]{3cm}
        \centering
        \includegraphics[width=3cm, height=3cm]{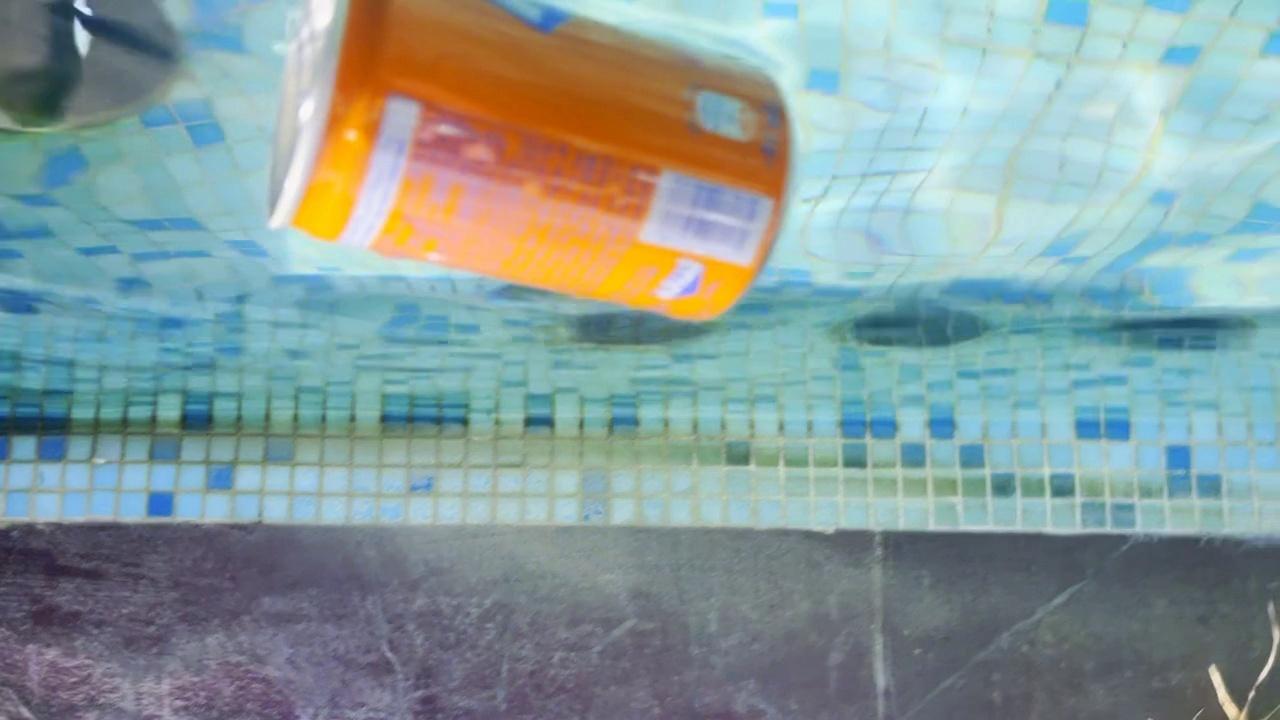} 
        \caption{Vertically flipped}
        \label{fig:image4}
    \end{subfigure}

    \caption{Examples of different image augmentations applied to a sample frame.}
    \label{fig:augmentation_grid}
\end{figure}

To improve the generalization of the models, several augmentation techniques were used that include brightness reduction, brightness gain, increased noise, increased color saturation, Gaussian blur, addition of different weather obstructions (clouds, fog, etc.), compression, mosaic effects, horizontal flipping, and vertical flipping seen in Fig \ref{fig:augmentation_grid}. The script to apply the augmentations utilises the libraries NumPy and OpenCV2 for image processing and pixel manipulation. These augmentations aim to make the model more adaptable and able to operate in circumstances its host may be in realistically, including operation when little to no sunlight is illuminating the object(s) meant to be detected, or in environments where noise, fog, or distortion to the host’s cameras may obstruct the view of the model. Attentive care was taken to ensure that none of the augmentations manipulated image labels resulting in inaccurate labelling and therefore faulty training. 
\vspace{0cm}

\subsection {Object Detection and Tracking}

\begin{figure}[H]
    \centering
    \includegraphics[width=1\linewidth]{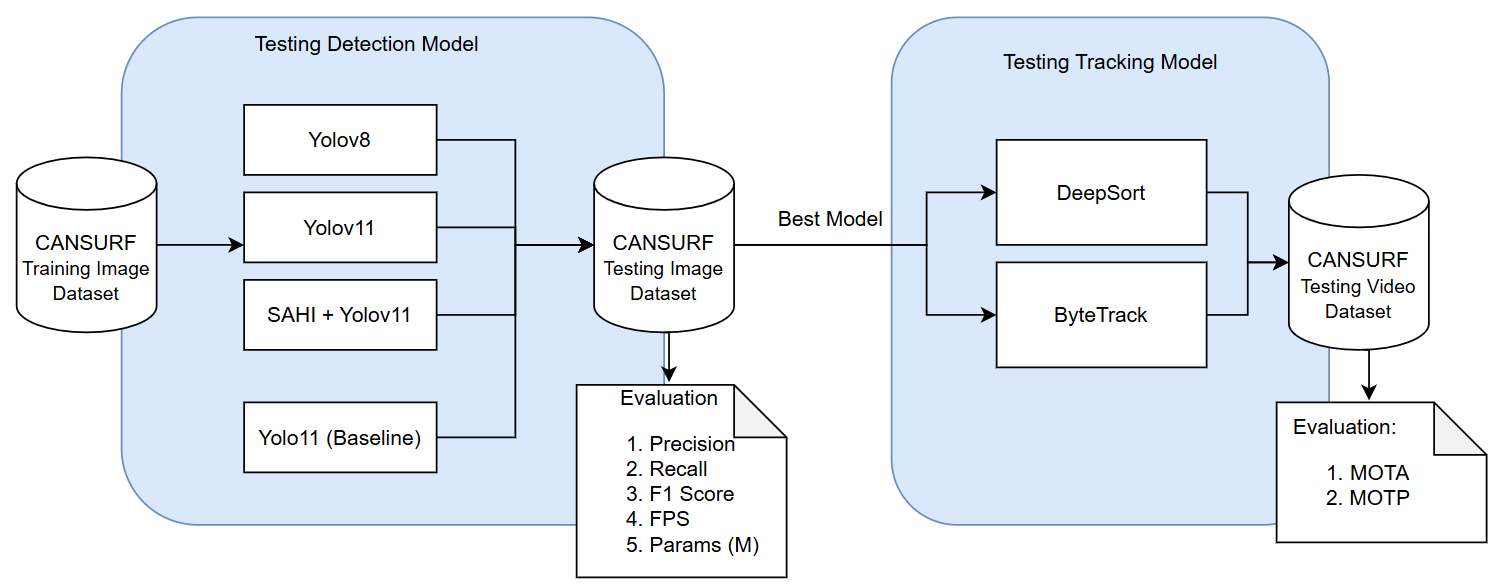}
    \caption{Pipeline of workflow.}
    \label{fig:pipeline_img}
\end{figure}

\subsubsection{Object Detection}
The core of the vision pipeline is a robust object detection model capable of accurately identifying floating cans. To select the optimal architecture for this task, a systematic benchmarking process was conducted on a curated dataset consisting of 900 images where cans occupy less than 5\% of the image frame. The goal of these tests is to identify a model that not only achieves the highest detection accuracy but also maintains the computational efficiency required for real-time deployment on the ASV. The used evaluation included three distinct approaches: established baseline models, zero-shot detection models, and an enhancing inference technique.
\vspace{-0.5cm}

\subsubsection{YOLOv8 and YOLOv11}
YOLOv8 and YOLOv11 (You Only Look Once) are leading state-of-the-art object detection models widely used by researchers and developers. These two architectures are known for their exceptional balance of speed and accuracy, making them the standard choice for many real-time applications. By benchmarking both,  their distinct architectural philosophies can be evaluated. YOLOv8 uses an anchor free design, which simplifies the detection pipeline and enhances performance on small objects. This makes it a strong baseline for detecting distant cans. On the other hand, YOLOv11 introduces more advanced components, notably the C2PSA (Parallel Split Attention) block. This attention mechanism is highly relevant to the intended use case, as it is designed to intensify the model's focus on significant object features while suppressing background noise, such as sun glare and water reflections, which are common sources of false positives in marine environments \cite{hidayatullah2025yolov8yolo11comprehensivearchitecture}.
\vspace{0cm}

\subsubsection{Zero-Shot Object Detection (ZSD)}
ZSDs are models that allow the identification of objects from categories not seen during training by aligning visual features with semantic information, allowing for generalization to novel classes at inference time\cite{Cao2025}. This capability is highly relevant for an autonomous system, which must identify a potentially limitless variety of marine debris without the need for constant re-annotation and retraining.
However, these models introduce significant challenges. Many state-of-the-art open-vocabulary detectors carry a high computational burden and present complex deployment challenges, making them less suitable for resource-constrained, real-time applications.
\vspace{0cm}

To evaluate this approach, two models will be benchmarked. Firstly, YOLO-World, a model designed for real-time open-vocabulary detection that allows high speed while detecting a flexible set of objects defined by text \cite{cheng2024yoloworldrealtimeopenvocabularyobject}. The second model for benchmarking is Grounding DINO, which combines a Transformer-based detector (DINO) with grounded pre-training, enabling it to accurately localize objects based on category names or referring expressions\cite{liu2024groundingdinomarryingdino}.
\vspace{-0.25cm}

\subsubsection{SAHI (Slicing Aided Hyper Inference)}
SAHI is an adaptive technique that can be applied to any object detection algorithm and improves a model’s performance in detecting smaller objects by by running inference on tiled image patches and merging the results. The subdivisions are scaled up to increase the coverage of a target object relative to its patch size This increase in relative object size is what enables the model to better detect the target. The SAHI framework was applied on top of the object detection models due to the small apparent size of cans in CANSURF, especially as the ASV is expected to be optimized to detect cans from afar. Given that many training images contained aluminum cans whose widths were \textless1\% of the image’s total width, the model has great potential to increase in performance using SAHI with little to no inference loss\cite{sahi_arxiv}.
\vspace{0cm}

\subsubsection{Detection Experimental and Training Setup}
The YOLOv8 and YOLOv11 models were trained for 150 epochs, with learning rate handled by the Ultralytics library. Early stopping was enabled as a precaution against overfitting. Models were trained for 150 epochs with early stopping on various GPUs from vast.ai (e.g., NVIDIA RTX 4070)\cite{vastai2024}. For benchmarking on the curated test dataset, all models were evaluated on an identical local machine equipped with an AMD RX 7600S GPU (8GB VRAM) to ensure a fair comparison of performance metrics. Furthermore, to realise the value of the specialised CANSURF dataset, a Baseline model was also trained and evaluated. This baseline consisted of a YOLOv11s model fine-tuned on generic can images (4,000 images) found online on the platform roboflow\cite{o_20bg_2_dataset}\cite{canettes-wjjyb_dataset}. This model was not exposed to any floating-can images, showcasing the perfomance challenges introduced by domain shift and the neccessity for a specialised domained dataset. The SAHI framework was also benchmarked in conjunction with the YOLOv11 architecture to assess its impact on small object detection, while ZSD models were considered conceptually but not trained due to their computational requirements and architectural complexity. For this reason, the ZSD models were evaluated in an open-vocabulary environment to compare against the closed-set tuned YOLO models trained on CANSURF.
\vspace{-0.25cm}

\subsubsection{Object Tracking}
Two multi-object tracking (MOT) algorithms were evaluated: DeepSORT and ByteTrack. Both methods were chosen because of their suitability for tracking multiple objects with an environment of erratic motion, where cans tend to drift unpredictably on the water’s surface.  

\textbf{DeepSORT} extends the baseline SORT algorithm by incorporating a deep appearance descriptor alongside the Kalman filter and Hungarian algorithm for data association. This appearance-based embedding is particularly beneficial for the intended use case, as it helps maintain consistent identities even when multiple cans overlap or partially occlude each in video streams. 

\textbf{ByteTrack}, on the other hand, introduces a high-recall association mechanism that leverages both high-confidence and low-confidence detections for tracking. This is advantageous in scenarios where water reflections, ripples, or lighting conditions temporarily reduce detector confidence, since ByteTrack can still propagate object trajectories without identity switches. 
\vspace{-0.25cm}

\subsubsection{Tracking Experimental Setup}
The performances of DeepSORT and ByteTrack were evaluated using YOLOv11 as the selected object detection model, as illustrated in Figure~\ref{fig:3}. Evaluation follows standard MOT benchmarks, two primary metrics were employed: Multiple Object Tracking Accuracy (MOTA) and Multiple Object Tracking Precision (MOTP).

MOTA finds the tracking performance with three sources of errors: false positives (FP), false negatives (FN), and identity switches (IDSW), defined as:

\begin{equation}
    \text{MOTA} = 1 - \frac{FN + FP + IDSW}{GT}
\end{equation}

where GT is the total number of ground-truth detections. A higher MOTA value signifies improved overall tracking accuracy- tracker’s capability to maintain object identities consistently across frames.

MOTP, in contrast, measures the spatial precision of the tracker by quantifying the average localization error between predicted and ground-truth bounding boxes. It is defined as:

\begin{equation}
    \text{MOTP} = \frac{\sum_{i,t} d_{i}^{t}}{\sum_{t} c_{t}}
\end{equation}

where $d_{i}^{t}$ represents the distance between matched detection and GT pairs at time $t$, and $c_{t}$. A higher MOTP value indicates superior localization accuracy..

Together, MOTA and MOTP provide a rigorous evaluation framework for tracking, when paired with YOLOv11 model, capturing the \textit{temporal consistency} and \textit{spatial precision} in tracking cans floating on water surface across video frames. 
\label{sec:meth}
\section{Results and Analysis}
\label{sec:raa}

\subsection{Benchmarking Object detection models}
To ensure a thorough and fair comparison, the models were evaluated using a set of standard performance metrics. These metrics provide insight into different aspects of a model's performance, from accuracy to computational efficiency. The results for all benchmarked models on the curated test dataset are summarized in Table \ref{tab:benchmark_results}.

\subsubsection{Validating the CANSURF Dataset}
The baseline YOLOv11s, trained on generic images, performed poorly with a low F1-score of 0.07. This shows a significant domain gap between models trained on standard can images compared to the unique marine challenges presented by the floating can images in the CANSURF dataset.

In contrast, models trained on CANSURF improved performance by about 12x. YOLOv11s achieves a superior F1-Score of 0.90. While YOLOv8s retains a higher FPS, the 133.5 FPS of YOLOv11s is more sufficient for real-time operation. The minor FPS difference may be attributed to specific optimisations in the Ultralytics library rather than a fundamental architectural advantage. Given its higher accuracy and smaller size (9.4M parameters), YOLOv11s stands out as the more efficient and effective model.

\subsubsection{Zero-shot models and SAHI}
\begin{figure}[h!]
\centering
\includegraphics[width=1\linewidth]{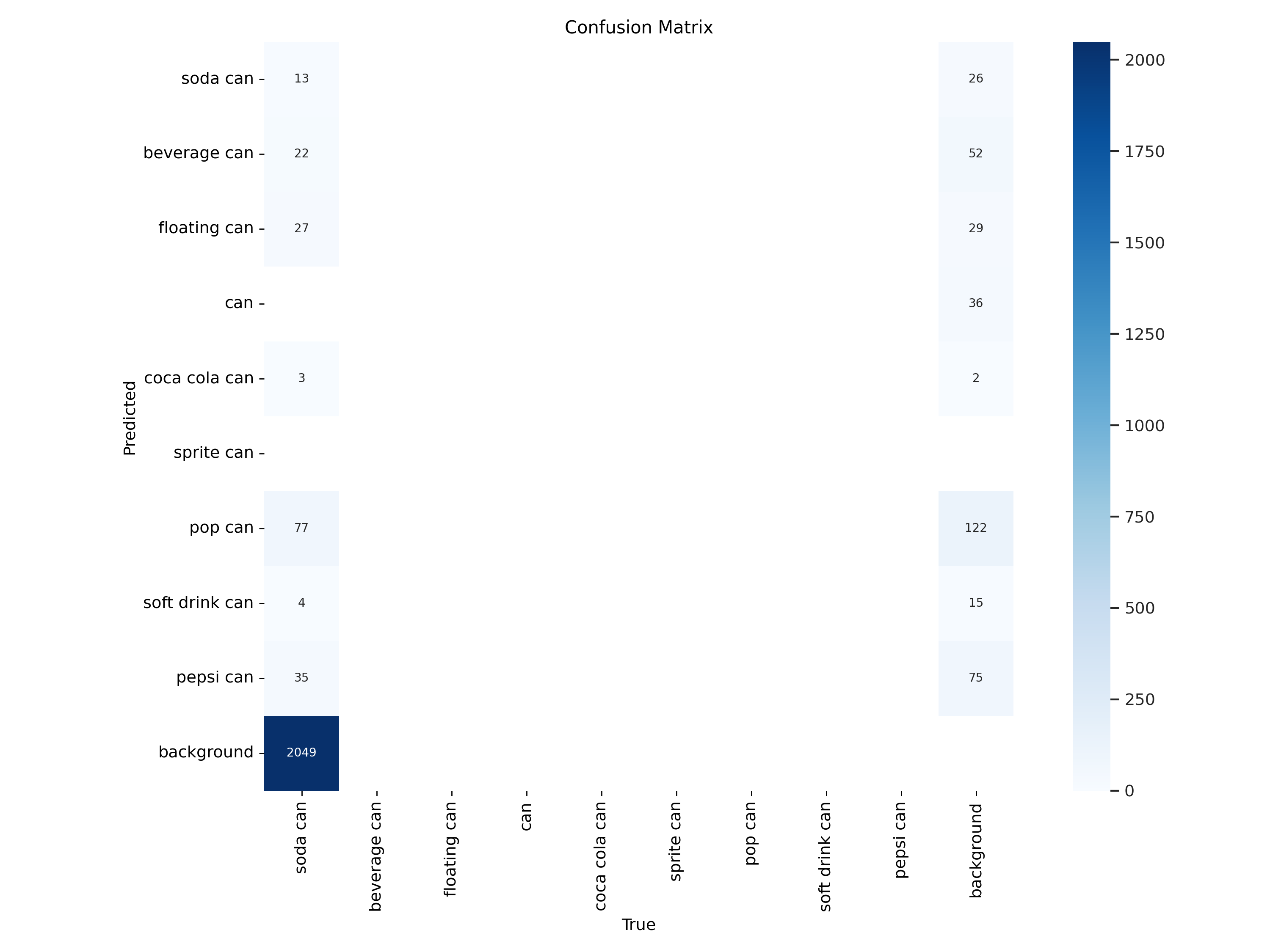}
\caption{YOLO-World's multi-class confusion matrix.}
\label{fig:yoloworld_cm}
\end{figure}

The Zero-shot models were tested to assess their potential for detecting multiple types of debris without specific training. YOLO-World showed some promise but was ultimately limited, as illustrated by its multi-class confusion matrix in Figure \ref{fig:yoloworld_cm}. By prompting the model with various synonyms for a can (e.g., 'soda can', 'beverage can'), its recall could be marginally improved by approximately 5\% , but the model still confused many objects and background elements. Grounding DINO performed poorly, with its low FPS of 1.9 confirming that its architecture is too computationally intensive for edge deployment. These results suggest that while Zero-Shot Detection is a promising field for future multi-class debris detection, current models lack the specialized accuracy required for this mission.

On the other hand, the application of SAHI to YOLOv11s yielded an increase in Recall to 0.93. However, this came at a severe cost to Precision (0.67) and a drastic reduction in speed to 5.8 FPS. The unjustifiably slow inference and high rate of false positives make this approach require more hardware power for autonomous vehicles.

\begin{table}[h]
\centering
\caption{Benchmarking Results for Object Detection Models on the curated test dataset}
\label{tab:benchmark_results}
\begin{threeparttable}
\setlength{\tabcolsep}{2pt} 
\footnotesize  
\begin{tabular}{lccccc}
\toprule
\textbf{Model} & \textbf{Precision} & \textbf{Recall} & \textbf{F1-score} & \textbf{FPS} & \textbf{Params (M)} \\
\midrule
YOLOv11s (Baseline) & 0.31 & 0.04 & 0.07 & 133.5 & 9.4 \\
\midrule
YOLOv8s (CANSURF)           & 0.89 & 0.89 & 0.89 & \textbf{155.3} & 11.2 \\
YOLOv11s (CANSURF)          & \textbf{0.90} & 0.90 & \textbf{0.90} & 133.5 & 9.4 \\
SAHI +
\\ YOLOv11s (CANSURF)   & 0.67 & \textbf{0.93} & 0.78 & 5.8   & 9.4\tnote{*} \\
\midrule
YOLO-World                  & 0.46 & 0.20 & 0.28 & 75.9  & 179 \\
Grounding DINO              & 0.09 & 0.20 & 0.12 & 1.9   & 172  \\
\bottomrule
\end{tabular}
\begin{tablenotes}
    \item[*] The parameter count is identical to the base YOLOv11s model as SAHI is an inference technique, not a change to the model architecture.
\end{tablenotes}
\end{threeparttable}
\end{table}

\vspace{0cm}
\subsection{Benchmarking DeepSORT and ByteTrack}

The quantitative results of DeepSORT and ByteTrack with YOLOv11 (Y11) and YOLOv11 + SAHI (Y11 SAHI) were evaluated on a single video lasting 2 minutes (3,600 frames at 30 FPS). Table~\ref{tab:results} reports the benchmarking results.

\begin{table}[h]
\centering
\caption{Benchmarking results on a custom dataset with different detectors and trackers.}
\label{tab:results}
\setlength{\tabcolsep}{2pt}
\begin{tabular}{lcccccccc}
\hline
\textbf{Detector} & \textbf{Tracker} & \textbf{MOTA} $\uparrow$ & \textbf{MOTP} $\uparrow$ & \textbf{FP} $\downarrow$ & \textbf{FN} $\downarrow$ & \textbf{IDs} $\downarrow$ & \textbf{FPS} $\uparrow$ \\ \hline
Y11           & DS   & 0.32 & 0.58 & 6 & 20  & 82 & 51.6 \\
Y11           & BT  & 0.535 & 0.70 & 4  & 35  & 36  & 50.0 \\
Y11 + SAHI    & DS   & 0.39 & 0.57 & 963 & 22  & 139 & 50.0 \\
Y11 + SAHI    & BT  & 0.437 & 0.61 & 335  & 11 & 58  & 45.0 \\ \hline
\end{tabular}
\end{table}
\vspace{0cm}
In evaluation, DeepSORT achieves reasonable bounding box accuracy but struggles with consistent IDs, and adding SAHI increases false positives and ID switches without improving overall tracking. ByteTrack provides more stable tracking with fewer ID switches and higher MOTA, and SAHI further reduces ID switches at the cost of additional false positives. Overall, YOLOv11 + SAHI + ByteTrack and YOLOv11 + ByteTrack offer the most reliable identity tracking of cans on water surface. However the former is more computationally heavy. 
\section{Conclusion}
\label{sec:conc}

This paper presented a surface-level, ASV-viewpoint dataset focused on aluminum cans floating on water and evaluated a detection–tracking pipeline for autonomous clean-up. Based on the experimentation, the strongest detector–tracker pair was YOLOv11 and ByteTrack, which achieved higher MOTA and MOTP, fewer ID switches, and competitive FPS compared to DeepSORT, indicating that ByteTrack provides more stable trajectories on top of a high-quality detector.
However, for the mission profile of reliable can detection to collect as many cans as possible, YOLOv11 + SAHI was chosen as the detection mode for its higher recall (.93) to avoid leaving the least amount of cans polluting its host water body. YOLOv11s trained on CANSURF still proves to be a very reliable option (.9 Precision) for general purpose detection of near distant cans.

To our knowledge, there is currently no other open dataset that targets aluminum cans on water, making this contribution a useful benchmark for the community. A key limitation of this dataset is the lack of testing and benchmarking under dynamic weather conditions such as rain, heavy wind, or highly choppy water. Another  limitation is that the YOLOv11+SAHI combination yielded lower precision in testing due to the prevalence of full-context (wide-FOV) frames in which cans are shown without obstruction, leading to the tiling done by SAHI producing partial can views which may not be recognised as cans, which can be mitigated by adding more far-distance examples, varied viewpoints, and non full-context images. Future work will expand the dataset to adverse weather and diverse locations, include longer video sequences for temporal benchmarking, and reassess detector–tracker trade-offs (and tiling-based inference) under those conditions and on varied edge-compute platforms.
\section*{Acknowledgment}
The authors would like to thank Dr.~Claudio Zito for his thorough review and insightful feedback, which substantially improved the clarity and quality of this paper.
\section*{Dataset Access}
The CANSURF dataset is available for research purposes on Zenodo and Github: 

\begin{itemize}
    \item Zenodo Link: \url{https://doi.org/10.5281/zenodo.20100657}
    \item Github Link: \url{https://github.com/ZaidAljundiHW2/CANSURF}
\end{itemize}




\bibliographystyle{IEEEtran}
\bibliography{references}

\end{document}